\def\paperID{10992}
\def\confName{CVPR}
\def\confYear{2024}

\def\authorBlock{
    ChengHuang\thanks{Equal contribution} ,
    Shoudong Han\footnotemark[1]  \thanks{Corresponding author},
    Mengyu He ,
    Wenbo Zheng ,
    Yuhao Wei
    \\
    Huazhong University of Science and Technology \\
    {\tt\small \{chenghuang7, shoudonghan\}@hust.edu.cn}
}

\newif\ifreview 
\newif\ifarxiv \newcommand{\arxiv}{\arxivtrue}
\newif\ifcamera 
\newif\ifrebuttal 

\arxiv 

\pdfoutput=1
\documentclass[10pt,twocolumn,letterpaper]{article}
\ifreview \usepackage[review]{cvpr} \fi
\ifarxiv \usepackage[pagenumbers]{cvpr} \fi
\ifrebuttal \usepackage[rebuttal]{cvpr} \fi
\ifcamera \usepackage{cvpr} \fi

\usepackage{graphicx}	
\usepackage{amsmath}	
\usepackage{amssymb}	
\usepackage{booktabs}
\usepackage{times}
\usepackage{microtype}
\usepackage{epsfig}
\usepackage[table,xcdraw,dvipsnames]{xcolor}
\usepackage{caption}
\usepackage{float}
\usepackage{placeins}
\usepackage{color, colortbl}
\usepackage{stfloats}
\usepackage{enumitem}
\usepackage{tabularx}
\usepackage{xstring}
\usepackage{multirow}
\usepackage{xspace}
\usepackage{url}
\usepackage{subcaption}
\usepackage{xcolor}
\usepackage[hang,flushmargin]{footmisc}
\usepackage{indentfirst}
\usepackage{colortbl}  
\usepackage{xcolor}
\usepackage{soul}
\usepackage{array}
\usepackage{makecell}

\ifcamera \usepackage[accsupp]{axessibility} \fi

% ...

\ifarxiv  \fi

\newcommand{\R}[1]{{%
    \textbf{%
        \ifstrequal{#1}{1}{\textcolor{red}{R#1}}{%
        \ifstrequal{#1}{2}{\textcolor{blue}{R#1}}{%
        \ifstrequal{#1}{3}{\textcolor{magenta}{R#1}}{%
        \ifstrequal{#1}{4}{\textcolor{teal}{R#1}}{%
                           \textcolor{cyan}{R#1}%
        }}}}%
    }%
}}

\usepackage{xr-hyper}

\makeatletter
\newcommand*{\addFileDependency}[1]{
  \typeout{(#1)}
  \@addtofilelist{#1}
  \IfFileExists{#1}{}{\typeout{No file #1.}}
}

\makeatother

\definecolor{cvprblue}{rgb}{0.21,0.49,0.74}
\usepackage[pagebackref,breaklinks,colorlinks,citecolor=cvprblue]{hyperref}
\usepackage[capitalize]{cleveref}
\crefname{section}{Sec.}{Secs.}
\crefname{table}{Table}{Tables}
\crefname{figure}{Fig.}{Figs.}

\begin{document}
\title{DeconfuseTrack: Dealing with Confusion for Multi-Object Tracking}
\author{\authorBlock}

\maketitle

\maketitle

\begin{abstract}
Accurate data association is crucial in reducing confusion, such as ID switches and assignment errors, in multi-object tracking (MOT). However, existing advanced methods often overlook the diversity among trajectories and the ambiguity and conflicts present in motion and appearance cues, leading to confusion among detections, trajectories, and associations when performing simple global data association. To address this issue, we propose a simple, versatile, and highly interpretable data association approach called Decomposed Data Association (DDA). DDA decomposes the traditional association problem into multiple sub-problems using a series of non-learning-based modules and selectively addresses the confusion in each sub-problem by incorporating targeted exploitation of new cues. Additionally, we introduce Occlusion-aware Non-Maximum Suppression (ONMS) to retain more occluded detections, thereby increasing opportunities for association with trajectories and indirectly reducing the confusion caused by missed detections. Finally, based on DDA and ONMS, we design a powerful multi-object tracker named DeconfuseTrack, specifically focused on resolving confusion in MOT. Extensive experiments conducted on the MOT17 and MOT20 datasets demonstrate that our proposed DDA and ONMS significantly enhance the performance of several popular trackers. Moreover, DeconfuseTrack achieves state-of-the-art performance on the MOT17 and MOT20 test sets, significantly outperforms the baseline tracker ByteTrack in metrics such as HOTA, IDF1, AssA. This validates that our tracking design effectively reduces confusion caused by simple global association. 
\end{abstract}
\section{Introduction}
\label{sec:intro}

Multi-object tracking (MOT) is a crucial task in the field of computer vision with extensive applications, including video surveillance \cite{1}, autonomous driving \cite{2}, and human-computer interaction \cite{3}. The goal of MOT is to track multiple objects of interest simultaneously in a video sequence. Despite significant advancements in this field, MOT still faces several challenges, such as occlusion, appearance variations, and complex interactions between objects.

In recent years, most MOT methods \cite{4,5,6,7,8} adopt the tracking-by-detection paradigm. In this paradigm, data association plays a crucial role in MOT by establishing correspondences between tracking trajectories and detection results. To improve the accuracy of data association, many methods introduce additional cues to complement motion cues, including appearance features \cite{5,9}, motion direction \cite{8,10}, confidence scores \cite{7,11}, depth information \cite{12,13}, and natural language cues \cite{14}. These additional cues are shown to alleviate issues caused by the ambiguity of motion cues or motion estimation errors to some extent. Furthermore, some methods \cite{7,15,16} divide the data association process into multiple stages, assigning priorities to trajectories and detection results through incremental matching, thereby reducing confusion during association.

\begin{figure}[tp]
    \centering
    \includegraphics[width=\linewidth]{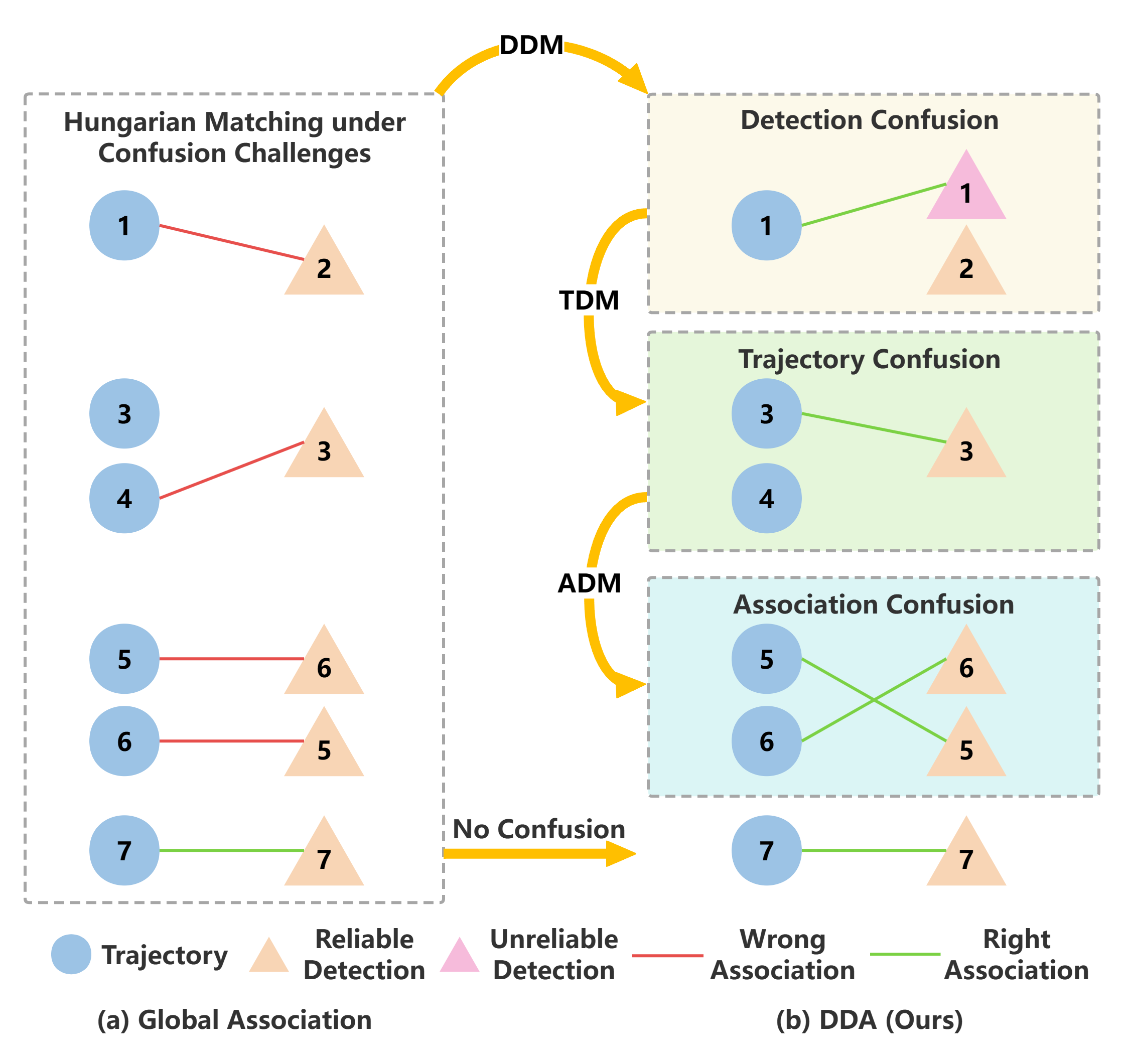}
    \caption{Comparing different data association methods. (a) Global Association. (b) Decomposed Data Association (Ours)}
    \label{fig:1}
\end{figure}

However, current MOT methods still have some limitations in terms of data association, primarily in two aspects. Firstly, many state-of-the-art methods \cite{6,8,9} treat data association as a global optimization problem, considering the assignment between all tracking trajectories and detection results as a single optimization task. However, such holistic association approaches may retain numerous confusions, leading to a degradation in tracking performance. As shown in \cref{fig:1}(a), global association often leads to numerous confusions. For instance, Trajectory 1 is incorrectly associated with Detection 2, which should be initialized as a new track. In another case, Trajectory 4 is mistakenly matched with Detection 3. Additionally, Trajectory 5 and 6 experience ID switches due to their close proximity. Secondly, although many methods consider using multiple cues to complement motion information, these methods \cite{6,9} either simply linearly weight the multiple cues or utilize heuristic rules \cite{10,17} for fusion. This approach, on one hand, makes the tracker highly sensitive to fusion hyperparameters, and on the other hand, may introduce uncertainties from the new cues that could potentially interfere with the accuracy of motion cues.

To address these issues, we propose a simple yet effective multi-object tracking method called DeconfuseTrack, which aims to tackle confusion in data association. Firstly, we suggest a more detailed consideration of the data association problem in multi-object tracking. We decompose the global association problem into several sub-problems, including the association between a single trajectory and multiple detections, the association between multiple trajectories and a single detection, and the association between multiple trajectories and multiple detections. By considering these sub-problems more thoroughly, we can minimize erroneous association assignments. When utilizing appearance cues, we adopt a decoupling strategy, only supplementing with appearance cues when the discriminative power of motion cues is insufficient in the sub-problems. The use of appearance cues is constrained within a certain range to minimize interference with motion cues. As shown in \cref{fig:1}(b), Through Detection Disambiguation Module (DDM), we identify Detection 1 that, although unreliable, is a better fit for Trajectory 1, freeing up Detection 2. With Trajectory Disambiguation Module (TDM), we make the correct selection between Trajectory 3 and Trajectory 4 for Detection 3. Through Association Disambiguation Module (ADM), we avoid association confusion between Trajectory 5 and Trajectory 6. Secondly, to enhance detection performance and mitigate confusion caused by missed detections, we design Occlusion-aware Non-Maximum Suppression (ONMS) to retain more occluded detection boxes for association. Extensive experimental results demonstrate that our proposed DeconfuseTrack method outperforms state-of-the-art methods on two widely adopted benchmark datasets, MOT17 \cite{18} and MOT20 \cite{19}.

Our work has made the following main contributions:

\begin{itemize}
  \item We design a novel plug-and-play data association method called Decomposed Data Association (DDA). It decomposes the traditional global data association into a series of sub-problems and handles them step by step, reducing the confusion in the assignment stage of MOT matching.
  \item We propose ONMS, a method that preserves more occluded detections for data association in the post-detection processing stage. This approach has the potential to reduce the occurrence of confusion during association.
  \item By combining DDA and ONMS, we propose a simple yet powerful multi-object tracker named DeconfuseTrack, to address the challenges of confusion in MOT. 
\end{itemize}

\section{Related Work}
\label{sec:related}
\noindent \textbf{Tracking-by-Detection.} Among the frameworks utilized in MOT, the tracking-by-detection paradigm stands out as the earliest and most widely embraced approach. It aims to detect objects in video frames using an object detector and then connect the objects across frames using data association methods. The performance of detection plays a crucial role in improving tracking performance. Therefore, some methods choose to use better detectors to obtain improved detection results. For example, the MOT17 dataset \cite{18} uses DPM \cite{23}, Faster-RCNN \cite{24}, and SDP \cite{25}. CenterNet \cite{29} is adopted by many methods \cite{9,26,27,28} due to its simplicity and ease of use. YOLOX \cite{30} has become the choice of most MOT methods \cite{7,8,6,12,22,31,32} due to its powerful detection performance. Another category of methods focus on improving the accuracy of object motion prediction to better associate objects across frames. For instance, many methods \cite{4,5,7,8,9,33} employ Kalman filters \cite{20} for motion prediction. Some methods \cite{6,17,34} consider using camera motion compensation to assist in object motion prediction. A few methods \cite{12,22,35} utilize learnable models for motion prediction. In addition, appearance modeling is crucial for improving object discrimination. Some methods \cite{5,6,17,36} use independent Re-Identification (ReID) models to extract appearance features of the targets. Other methods \cite{37,9,38,39} incorporate the ReID task as a branch of the detector, enabling a single model to simultaneously perform detection and embed target features.

Our approach follows the popular tracking-by-detection paradigm and maintains the same configuration as the popular methods \cite{7,8,17} in terms of motion prediction, ReID, and other aspects. However, we observe that the majority of methods simply employ NMS to filter out duplicate detections, resulting in the loss of many detected occluded objects and underutilization of the detector's performance. Therefore, we chose to use a simple ONMS technique to retain as many detections as possible.

\noindent \textbf{Data Association.} Data association is a crucial module in multi-object trackers, aiming to accurately assign tracking trajectories to detections. Data association methods in MOT can be traced back to the radar domain, such as JPDA \cite{40, 41} and MHT \cite{42}. However, these algorithms have high computational complexity and require prior assumptions about the number of targets, making them less commonly adopted in modern visual MOT. Many visual MOT data association methods are based on SORT \cite{4}, which utilizes a simple Hungarian matching \cite{21} to assign detections to tracks. DeepSORT \cite{5} introduces a cascaded matching approach that categorizes tracks into ``confirmed'' and ``tentative'' states. It prioritizes matching the confirmed tracks before considering the tentative ones, reducing identity switches during tracking. ByteTrack \cite{7} proposes a multi-stage association method that first associates tracks with high-scoring detections in the first stage, and then associates the remaining tracks with low-scoring detections in the second stage. By utilizing more detections, it significantly reduces false negatives. Another category of methods achieve implicit data association using learnable models. For example, some methods \cite{43, 44, 45, 46} employ Graph Neural Networks (GNN) to model similarity and data association. Another set of methods \cite{47, 48, 49, 50}  utilize the query mechanism of Transformers \cite{51} for association.

Although different trackers employ various data association methods, they typically adopt a global association approach, often overlooking the individual characteristics of trajectories and detections while neglecting the ambiguity in the clues. In contrast, our method utilizes a decomposed data association technique, which tackles the ambiguity from a more granular perspective. Moreover, we don't incorporate any learnable modules, striking a balance between speed and interpretability.

\section{Method}
\label{sec:method}
\begin{figure*}[tp]
    \centering
    \includegraphics[width=\linewidth]{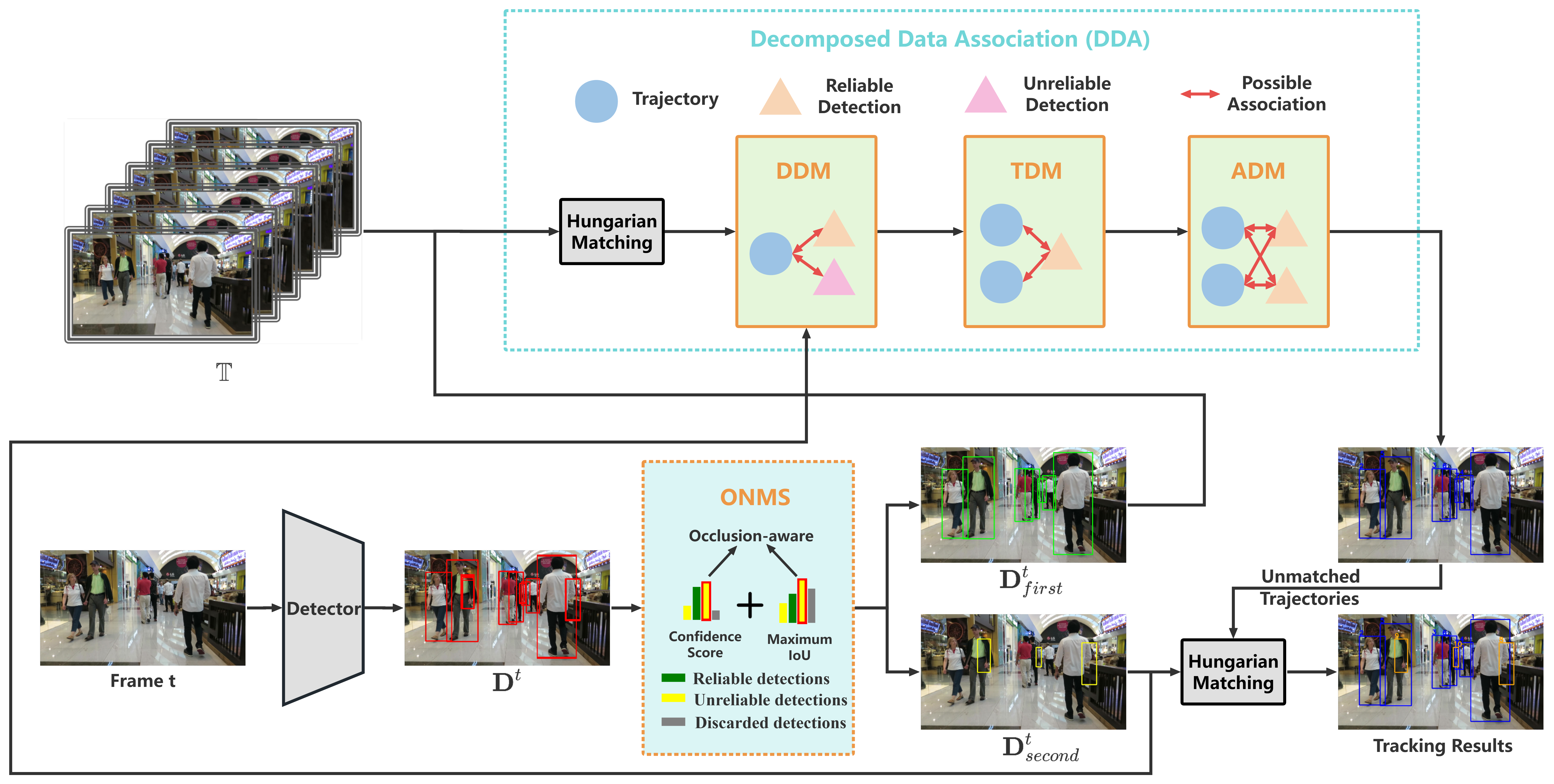}
    \caption{The overall pipeline of DeconfuseTrack. (1) Utilizing a detector to obtain the detection results for the current frame. (2) Employing ONMS to separate the detection results into reliable and unreliable detections. (3) Performing the first association using DDA. (4) Conducting the second association using unreliable detections and unassociated trajectories.}
    \label{fig:2}
\end{figure*}

\subsection{Notation}
Our method follows the popular tracking-by-detection paradigm. Firstly, we utilize a detector to obtain detection results for each frame. The detection results for frame $t$ can be represented as $\mathrm{D}^{t}=\{\mathrm{d} _{i}^{t}\mid {i}\in\{1,2,\cdots,N\}\}$, where $N$ is the number of detection boxes in the current frame. Each detection $\mathrm{d}_{i}^{t}\in\mathbb{R}^{5}$ can be represented as $\mathrm{d}_{i}^{t}=(x,y,w,h,c)$, where $(x,y)$ denotes the center coordinates of the bounding box, $w$ and $h$ represent the width and height of the bounding box, and $c$ is the confidence score of the detection. Trajectories can be represented as $\mathbb{T}=\{{\mathcal{T}}_{j}\mid j\in\{1,2,\cdots,M \}\}$, where $M$ is the total number of trajectories. Each trajectory is defined as ${\mathcal{T}}_{j}=\{\mathrm{o}^s\mid s\in\{t_{s},t_{s}+1,\cdots,t_{e}\}\}$, where $j$ is the identity of the trajectory, $t_{s}$ represents the initialization time of the trajectory, $t_{e}$ represents the termination time of the trajectory, and $\mathrm{o}^s=(x,y,w,h)$ represents the position of the trajectory at time $s$. We divide $\mathrm{D}^{t}$ into two categories like ByteTrack\cite{7}: reliable and unreliable, denoted as $\mathrm{D}^{t}_{fisrt}$  and $\mathrm{D}^{t}_{second}$ respectively. They are used for the first and second data associations. This will be explained in detail in \cref{sec:pnms}.

\subsection{Decomposed Data Association (DDA)}

To tackle the assignment problem and alleviate confusion in tracking, we propose the DDA method. For each frame, similar to popular approaches \cite{4,7,8}, we utilize the Kalman filter \cite{20} to obtain predicted positions ${\mathrm{L}}=\{{\mathrm{l}}_{j}=(x,y,w,h)\mid j\in\mathbb{T}\}$ for each trajectory in $\mathbb{T}$. The positional similarity between $\mathrm{d}_{i}$ and $\mathcal{T}_{j}$ is defined as the IoU between the detection bounding box and the predicted bounding box of the trajectory:

\begin{equation}
\operatorname{LocSim}(\mathrm{d}_{i},{\mathcal{T}}_{j})=\operatorname{IoU}(\mathrm{d}_{i},\mathrm{l}_{j})\,.
\end{equation}
Then use $\mathrm{D}^t_{first}$ and $\mathrm{L}$ to calculate the cost matrix $\mathrm{C}$:
\begin{equation}
    \mathrm{C}_{ij}=1-\mathrm{LocSim}(\mathrm{d}_{i},{\mathcal{T}}_{j})\,,\,\mathrm{d}_{i}\in \mathrm{D}_{first}^{t}\,,\,{\mathcal{T}}_{j}\in\mathbb{T}\,.
\end{equation}
Finally, we utilize the Hungarian algorithm \cite{21} to solve $\mathrm{C}$ and obtain the allocation result $\mathrm{P}=\{(\mathrm{d},\mathcal{T})\mid\mathrm{d}\in\mathrm{D}^{t}_{matched}\,,\,\mathcal{T}\in\mathbb{T}_{matched}\}$ where $\mathrm{D}^{t}_{matched}$ represents matched detections, and $\mathbb{T}_{matched}$ represents matched tracks. Previous methods would directly output the assignment result at this stage. However, this global association approach still retains some errors due to the confusion in the clues. To obtain more accurate assignment results, we aim to refine P and achieve a finer-grained allocation.

\noindent \textbf{Detection Disambiguation Module (DDM).} DDM aims to resolve the confusion between multiple detections and a single trajectory. However, during tracking, trajectories are usually more abundant than reliable detections, making it challenging to encounter situations where multiple reliable detections correspond to a single trajectory. If we were to consider unreliable detections as well, the number of detections would far exceed the number of trajectories. Nonetheless, blindly relying on unreliable detections would introduce numerous errors, as their appearance information is generally unreliable. Therefore, in DDM, we choose to solely utilize motion cues for deconfusion, as they are more reliable in this context.

 For the $j$-th assignment pair in $\mathrm{P}$, we identify the set of detection boxes that could potentially cause confusion with this assignment:
 \begin{align}
\mathrm{D}_{blur}^{j} &= \{\mathrm{d}_{i}\mid\operatorname{LocSim}(\mathrm{d}_{i},\mathcal{T} _j)-\operatorname{LocSim}(\mathrm{d}_{j},\mathcal{T} _j)
> \kappa\;,\;\nonumber \\&\mathrm{d}_{i}\in \mathrm{D} _{second}^{t},\mathrm{d}_{j}\in\mathrm{D}^{t}_{matched},\mathcal{T} _j\in\mathbb{T}_{matched} \}\,,
\end{align}
where $\kappa$ is the confusion reduction factor, this process is equivalent to finding potentially more suitable unreliable detections for each matched trajectory. Next, we define the assignment relationship $\mathrm{P}_{new}$, where the assignment pairs in $\mathrm{P}_{new}$ represent the deconfused assignment pairs resulting from our deconfusion process:
\begin{equation}
\begin{aligned}
&P_{new}=\{(\mathrm{d},\mathcal{T}_j)\mid\mathrm{d}=\operatorname*{argmax}_{d_i\in\mathrm{D}^j_{blur}}\operatorname{LocSim}(\mathrm{d}_i,\mathcal{T} _j)\\
&\operatorname{if~}\mathrm{D}_{blur}^{j}\not=\phi,\mathrm{d}_{j}\in\mathrm{D}^{t}_{matched},\mathcal{T} _j\in\mathbb{T}_{matched}\}\,.
\end{aligned}
\end{equation}
In the case of conflicts where the same unreliable detection may be selected by multiple trajectories, we retain only the assignment with a higher positional similarity. Finally, we move the unreliable detection boxes matched in $\mathrm{P}_{new}$ into the reliable detection boxes:
\begin{equation}
\begin{aligned}
{\mathrm{D} _{first}^{t}}^{'}&={\mathrm{D} _{first}^{t}}\cup\{\mathrm{d}\mid(\mathrm{d},\mathcal{T})\in{\mathrm{P}}_{new}\}\\  
{\mathrm{D} _{second}^{t}}^{'}&={\mathrm{D} _{second}^{t}}-\{\mathrm{d}\mid(\mathrm{d},\mathcal{T})\in{\mathrm{P}}_{new}\}\,.
\end{aligned}
\end{equation}
By increasing $\kappa$, we can ensure that the trajectories in $\mathrm{P}_{new}$ find much more suitable detections compared to those in the original $\mathrm{P}$. After obtaining $\mathrm{P}_{new}$, we potentially free up some reliable detection boxes in $\mathrm{P}$, as they are replaced by more appropriate unreliable detection boxes. However, these mismatched reliable detection boxes still have the possibility of being associated with unmatched trajectories. Therefore, in the final step, while ensuring the validity of the assignment relationship in $\mathrm{P}_{new}$, we perform a reassignment of $\mathbb{T}$ and ${\mathrm{D} _{first}^{t}}^{'}$ to obtain the new assignment relationship $\mathrm{P}_{ddm}$ after detection disambiguation.

\noindent \textbf{Trajectory Disambiguation Module (TDM).} Targets in the tracking process are prone to fragmentation due to occlusion, rapid motion, and other factors, leading to the formation of multiple trajectories. Additionally, erroneous initialization of false detections can also contribute to the increase in the number of trajectories. As a result, there is a challenge of matching multiple trajectories to a single detection. Furthermore, factors such as camera motion, long-term target absence, and inaccurate detector localization contribute to the ambiguity of the predicted positions $\mathrm{L}$. Relying solely on motion cues can lead to confusion. In light of these challenges, we choose to incorporate appearance cues to alleviate the confusion between trajectories and compensate for the limitations of motion information.

First, we identify all unmatched trajectories $\mathbb{T}_{lost}=\mathbb{T}-\mathbb{T}_{matched}$ in the current frame. These trajectories may have been erroneously rejected due to the ambiguity of motion cues. For the $j$-th assignment pair $(\mathrm{d}_j,\mathcal{T}_j)$ in $\mathrm{P}$, we then identify the set of trajectories that may cause confusion with this assignment:

\begin{align}
&\mathbb{T}_{blur}^{j}=\{{\mathcal{T}}_{i}\mid\operatorname{LocSim}(\mathrm{d}_{j},{\mathcal{T}}_{j})-\operatorname{LocSim}(\mathrm{d}_{j},{\mathcal{T}}_{i}) \nonumber <\kappa\,,\,\\&{\mathcal{T}}_{i}\in\mathbb{T}_{lost},\mathrm{d}_{j}\in\mathrm{D}^{t}_{matched},\mathcal{T} _j\in\mathbb{T}_{matched}\}\cup{\mathcal{T}}_{\mathrm{j}}\,.
\end{align}
\
The parameter $\kappa$ represents the confusion reduction factor, which is used to adjust the degree of confusion reduction. A larger value of $\kappa$ indicates a higher level of distrust in the positional cues. Next, we employ an appearance model to obtain more accurate assignments. The appearance embedding of trajectory $\mathcal{T}$ is denoted as $f_\mathcal{T}$ and the appearance embedding of detection $\mathrm{d}$ is denoted as $f_\mathrm{d}$. For each set of confused trajectories $\mathbb{T}_{blur}^{j}$, we select the trajectory with the closest appearance distance to $f_{\mathrm{d}_j}$:
\begin{equation}
\mathcal{T}_{best}^j=\operatorname*{argmin}_{\mathcal{T} \in\mathbb{T}_{blur}^j}\operatorname{CosDist}(f_{\mathrm{d}_{j}},f_\mathcal{T})\,,
\end{equation}
 where $\operatorname{CosDist}(\cdot)$ represents the calculation of cosine distance between two vectors. $\mathcal{T}_{best}^j$ can also refer to $\mathcal{T}_{j}$ itself. In the case of conflicts where a single trajectory may be selected as $\mathcal{T}_{best}^j$ by multiple detections, we only retain the assignment with the smaller cosine distance. Finally, we replace $\mathcal{T}_{j}$ in the original assignment pair with $\mathcal{T}_{best}^j$ to obtain the new assignment relationship after trajectory disambiguation:
 \begin{equation}
 \mathrm{P}_{tdm}=\{(\mathrm{d}_{j},\mathcal{T}_{best}^{j})\mid\mathrm{d}_{j}\in\mathrm{D}^{t}_{matched}\}.
 \end{equation}

 \noindent \textbf{Association Disambiguation Module (ADM).} During the tracking process, there can also be cases of target occlusion, target intersection, and other scenarios where we encounter confusion in associating multiple detections with multiple trajectories. For simplicity, we address the confusion between two detections and two trajectories at a time. Cases involving multiple-to-multiple associations can be decomposed into several two-to-two problems for resolution.

 First, for any two distinct assignments in $\mathrm{P}$, we use the coefficient of variation to quantify the confusion between them in terms of positional cues:

\begin{align}
&\operatorname{Cv}(i,j) = \frac{\operatorname{Std}\left(\left\{\operatorname{LocSim}\left(\mathrm{d}_{k_1}, \mathcal{T}_{k_2}\right) \mid k_1,k_2 \in\{i, j\}\right\}\right)}{\operatorname{Mean}\left(\left\{\operatorname{LocSim}\left(\mathrm{d}_{k_1}, \mathcal{T}_{k_2}\right) \mid k_1,k_2 \in\{i, j\}\right\}\right)},\nonumber\\
&\mathrm{d}_i, \mathrm{d}_j \in \mathrm{D}^{t}_{matched},\mathcal{T}_i, \mathcal{T}_j \in \mathbb{T}_{matched}, i \neq j \,.
\end{align}
When the coefficient of variation is small, it indicates that there is little difference in positional cues between the assignment pairs. As mentioned in TDM, positional cues are inherently ambiguous, so a small coefficient of variation implies a strong level of confusion between them. Conversely, a large coefficient of variation indicates a significant difference in positional cues between the assignment pair, suggesting a weak level of confusion. Next, we identify all the assignment pairs that exhibit strong confusion:
\begin{equation}
\mathrm{P}_{blur}=\left\{\left(\left(\mathrm{d}_i, \mathcal{T}_i\right),\left(\mathrm{d}_j, \mathcal{T}_j\right)\right) \mid \operatorname{Cv}(i, j)<\kappa\right\}\,,
\end{equation}
where $\kappa$ is the confusion reduction factor. Next, similar to TDM, we utilize appearance cues to resolve the positional confusion in $\mathrm{P}_{blur}$ and find more suitable assignment relationships. If the sum of the appearance distances of the assignment pairs in $\mathrm{P}_{blur}$ is smaller after resolving the cross-association, we consider the post-cross-association assignment pairs to be better and include them in the set $\mathrm{P}_{new}$:

\begin{align}
\mathrm{P}_{new}&=\{((\mathrm{d}_i, \mathcal{T}_j),(\mathrm{d}_j, \mathcal{T}_i)) \mid \operatorname{CosDist}(f_{\mathrm{d}_i}, f_{\mathcal{T}_j})\nonumber\\
&+\operatorname{CosDist}(f_{\mathrm{d}_j}, f_{\mathcal{T}_i}) < \operatorname{CosDist}(f_{\mathrm{d}_i}, f_{\mathcal{T}_i})\\
&+\operatorname{CosDist}(f_{\mathrm{d}_j}, f_{\mathcal{T}_j}),((\mathrm{d}_i, \mathcal{T}_i),(\mathrm{d}_j, \mathcal{T}_j)) \in \mathrm{P}_{blur}\}\nonumber\,.
\end{align}
During complex matching involving the cross-association of multiple detections and trajectories, conflicts can arise. To resolve these conflicts, we perform the Hungarian matching algorithm again using appearance cues to eliminate the conflicts in $\mathrm{P}_{new}$. Finally, we combine the revised assignment relationship $\mathrm{P}_{new}$ with the original set $\mathrm{P}$ to obtain the new assignment relationship $\mathrm{P_{adm}}$ after association disambiguation.

\noindent \textbf{Module Combination.} The DDM, TDM, and ADM modules are designed to take the assignment relationship $\mathrm{P}$ as input and generate a new assignment relationship $\mathrm{P}^{'}$. Therefore, these three modules can be combined in a serial manner to form the overall DDA. Considering that DDM modifies $\mathrm{D} _{first}^{t}$ and $\mathrm{D} _{second}^{t}$ to allow for more possibilities in subsequent modules, we prioritize using the DDM module. Since ADM deals with a larger scope than TDM, we place it at the end. All three modules share the confusion reduction factor $\kappa$ as a hyperparameter for robustness and simplicity. When $\kappa$ is increased, we consider more confusion cases in TDM and ADM, while fewer unreliable detections are considered in DDM. Thus, a larger value of $\kappa$ indicates less reliance on positional cues, whereas a smaller value indicates greater reliance on motion cues. The value of $\kappa$ can be flexibly adjusted based on the motion characteristics of the cameras and tracked objects in the dataset. Additionally, the DDA design does not include any learnable components. The appearance cues used for deconfusion can be obtained from any appearance model, making it easy to integrate the DDA onto other trackers in a flexible and convenient manner.

\begin{figure}[tp]
    \centering
    \includegraphics[width=\linewidth]{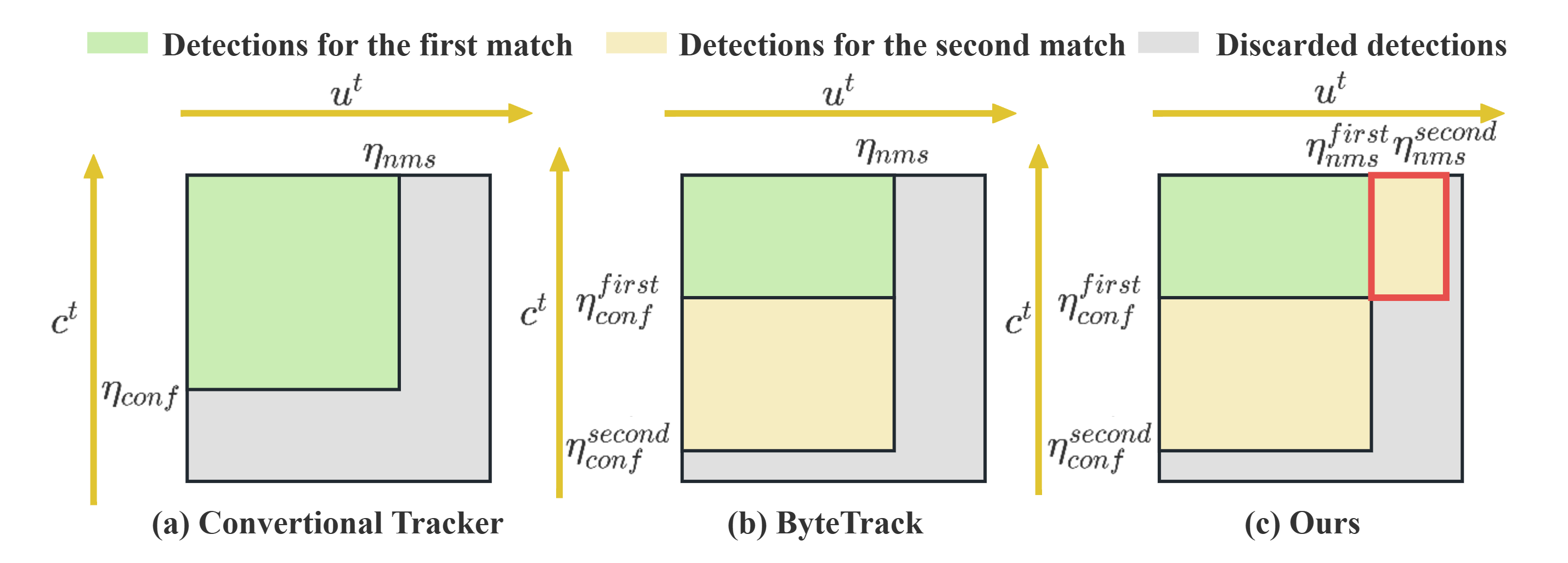}
    \caption{Comparing the post-processing approaches of different trackers: (a) Ordinary trackers use NMS and discard low-scoring detections. (b) ByteTrack also utilizes NMS but retains low-scoring detections. (c) Our method employs ONMS and retains occluded detectons.}
    \label{fig:3}
\end{figure}
\subsection{Occlusion-aware NMS (ONMS)}
\label{sec:pnms}
Improving the quality of detections can increase the success rate of data association and reduce incorrect associations. By reducing missed detections, more accurate location information can be obtained for the trajectories, reducing confusion caused by inaccurate motion predictions. Therefore, improving detections is crucial for enhancing MOT performance. 

We denote the confidence of $\mathrm{d}^t$ as $c^t$ and its maximum IoU with detections having higher confidence as $u^t$:
\begin{equation}
u_i^t=\max_{\mathrm{d} _j^t\in\{\mathrm{d}^t\mid c^t>c_i^t\,,\, \mathrm{d}^t\in \mathrm{D}^t\}}\operatorname{IoU}(\mathrm{d}_i^t,\mathrm{d}_j^t)\,.
\end{equation}
As shown in \cref{fig:3}(a), conventional trackers\cite{4,5,9,26,34} set a confidence threshold $\eta_{conf}$ and an NMS threshold $\eta_{nms}$ to retain only detections with $c^t$ higher than $\eta_{conf}$ and $u^t$ lower than $\eta_{nms}$ for a single global association. However, this approach mistakenly discards many correct detections. To address this issue, as illustrated in \cref{fig:3}(b), ByteTrack\cite{7} divides detections into two groups by setting two confidence thresholds $\eta_{conf}^{first}$ and $\eta_{conf}^{second}$, and performs two-stage associations to utilize more detections, significantly improving MOT performance. However, we believe that there is still room for improvement. In scenarios with dense target occlusion, we observe that detectors are not incapable of detecting heavily occluded objects. However, previous methods use a single NMS threshold $\eta_{nms}$ to post-process detection results, striking a balance between missed detections and false positives. Consequently, heavily occluded target boxes are discarded by NMS, even if they have high confidence scores. To address this limitation, as shown in \cref{fig:3}(c), we propose setting two NMS thresholds $\eta_{nms}^{first}$ and $\eta_{nms}^{second}$ to retain more detections for the data association stage:

\begin{align}
\mathrm{D}^t_{first}=&\{\mathrm{d}^t\mid \mathrm{d}^t \in \mathrm{D}^t,c^t\ge \eta_{conf}^{first},u^t\le \eta_{nms}^{first}\} \\
\mathrm{D}^t_{second}=&\{\mathrm{d}^t\mid \mathrm{d}^t \in \mathrm{D}^t,\eta_{conf}^{first}> c^t\ge \eta_{conf}^{second},u^t\le \eta_{nms}^{first}\} \nonumber\\ 
\cup &\{\mathrm{d}^t\mid \mathrm{d}^t \in \mathrm{D}^t,c^t\ge \eta_{conf}^{first},\eta_{nms}^{first}<u^t\le \eta_{nms}^{second}\} \nonumber
\end{align}

\subsection{DeconfuseTrack} 
By combining DDA and ONMS, we propose a tracker called DeconfuseTrack that focuses on addressing confusion in MOT. It adopts the popular tracking-by-detection architecture\cite{6,7,8} and utilizes ONMS to enhance the output of the detector, reducing confusion caused by insufficient detection capabilities. Additionally, DDA is employed for more precise data association, reducing confusion arising from ambiguous positional cues. The overall architecture is illustrated in \cref{fig:2}. For the first frame of each tracking video, we initialize $\mathbb{T}$ with $\mathrm{D}^t_{first}$. In the subsequent frames, we update $\mathbb{T}$ using $\mathrm{D}^t_{first}$ and $\mathrm{D}^t_{second}$. Unassociated detection boxes in $\mathrm{D}^t_{first}$ are added to $\mathbb{T}$ as newborn trajectories, while trajectories in $\mathbb{T}$ that have not been updated within a specified time are removed.
\section{Experiments}
\label{sec:experiments}
\subsection{Setting}
\noindent \textbf{Datasets.} We evaluate our DeconfuseTrack using the widely recognized MOT17 \cite{18} and MOT20 \cite{19} benchmarks, following the ``Private Detection" protocol. The MOT17 dataset consists of multiple multi-object tracking video sequences captured in natural scenes, providing high-quality annotation information. This dataset includes challenging scenarios such as camera motion and pedestrian occlusion, among others. The MOT20 dataset contains scenes with denser crowds, making it more prone to object confusion. Both the MOT17 and MOT20 datasets provide only training and testing sets. For ablation experiments, we follow the convention proposed in \cite{26}, where we use the first half of each video in the MOT17 training set for training and the second half for validation.

\noindent \textbf{Metrics.} We utilize widely accepted evaluation metrics, including the CLEAR metrics \cite{52}, IDF1 \cite{53}, and HOTA \cite{54} (including AssA, DetA). The MOTA, DetA primarily focus on the detection performance, while IDF1, AssA primarily assess the association performance. HOTA provides a balanced evaluation of both detection and tracking performance. DetA primarily reflects detection performance, which fluctuates only slightly around zero points in our experimental process.Therefore, we did not highlight DetA in our experiments.

\noindent \textbf{Implementation Details.} We implemented DeconfuseTrack within the MMTracking framework \cite{55} and selected ByteTrack \cite{7} as the baseline. To ensure a fair comparison, we adopted all the hyperparameter settings of ByteTrack and used the same YOLOX \cite{30} detector trained in ByteTrack. For the appearance model, we trained the SBS-50 model from FastReID \cite{56}  for 60 epochs on both MOT17 and MOT20 datasets. Regarding DDA, we selected confusion reduction factor $\kappa$ to 0.3. For ONMS, we set the thresholds $\eta_{nms}^{first}$ and $\eta_{nms}^{second}$ to 0.7 and 0.95, respectively.

\begin{table}[t]
\centering
\resizebox{\columnwidth}{!}{

\begin{tabular}{lcccccc}
    \toprule[2pt]
        Method & Venue & HOTA$\uparrow$  & IDF1$\uparrow$  & MOTA$\uparrow$  & AssA$\uparrow$  & DetA$\uparrow$  \\ \midrule
        \multicolumn{7}{l}{Learnable Matcher:}  \\ 
        MOTR\cite{48} & ECCV'22 & 57.8  & 68.6  & 73.4  & 55.7  & 60,3 \\ 
        MeMOT\cite{50} & CVPR'22 & 56.9  & 69.0  & 72.5  & 55.2  & - \\ 
        MOTRv2\cite{32} & CVPR'23 & 62.0  & 75.0  & 78.6  & 60.6  & 63.8 \\ 
        UTM\cite{46} & CVPR'23 & 64.0  & 78.7  & \textbf{81.8}  & - & - \\ 
        MeMOTR\cite{49} & ICCV'23 & 58.8  & 71.5  & 72.8  & 58.4  & 59.6 \\ 
        \multicolumn{7}{l}{Non-Learnable Matcher:} \\ 
        FairMOT\cite{9} & IJCV'21 & 59.3  & 72.3  & 73.7  & 58.0  & 60.9 \\ 
        QDTrack\cite{38} & CVPR'21 & 53.9  & 66.3  & 68.7  & 52.7  & 55.6 \\ 
        RelationTrack\cite{57} & TMM'22 & 61.0  & 74.7  & 73.8  & 61.5  & 60.6 \\ 
        MTracker\cite{MTracker} & ECCV'22 & -  & 75.9  & 77.3 & -  & - \\ 
        \rowcolor{pink!50}ByteTrack\cite{7} & ECCV'22  & 63.1  & 77.3  & {80.3}  & 62.0  & 64.5 \\ 
        \rowcolor{pink!50}QuoVadis\cite{12} & NeurIPS'22 & 63.1  & 77.7  & {80.3}  & 62.1  & {64.6} \\ 
        \rowcolor{pink!50}RTU++\cite{RTU++} & TIP'22 & 63.9  & 79.1  & {79.5}  & 63.7  & {64.5} \\ 
        \rowcolor{pink!50}SAT\cite{31} & ACM MM'22 & {64.4}  & {79.8}  & 80.0  & 64.4 & 64.8 \\ 
        \rowcolor{pink!50}C-BIOU\cite{CBIOU} & WACV'23  & 64.1  & 79.7  & 81.1  & {63.7}  & 64.8 \\ 
        \rowcolor{pink!50}StrongSORT++\cite{6} & TMM'23  & 64.4  & 79.5  & 79.6  & {64.4}  & 64.6 \\ 
        \rowcolor{pink!50}OC-SORT\cite{8} & CVPR'23 & 63.2  & 77.5  & 78.0  & 63.2  & - \\ 
        \rowcolor{pink!50}GHOST\cite{58} & CVPR'23 & 62.8  & 77.1  & 78.9  & - & - \\ 
        \hline
        \rowcolor{pink!50}\textbf{DeconfuseTrack} & - & \textbf{64.9}  & \textbf{80.6}  & 80.4  & \textbf{65.1}  & \textbf{65.0} \\ \bottomrule[2pt]
    \end{tabular}}
    \caption{Comparing with state-of-the-art methods on the MOT17 test set under the private detection protocol. The methods within the \sethlcolor{pink!50}\hl{pink} block utilize YOLOX \cite{30} as the detector. The best results are highlighted in \textbf{bold}.}
\vspace{-0.05in}
\label{tab:1}
\end{table}
\subsection{Comparison with the State-of-the-art Methods}
\noindent \textbf{MOT17.} \cref{tab:1} presents the performance of DeconfuseTrack on the MOT17 test dataset. Compared to the baseline  ByteTrack, our method shows significant improvements in association performance, with an increase of 1.8\% in HOTA, 3.3\% in IDF1, and 3.1\% in AssA. These results indicate that our proposed DDA and ONMS methods serve as strong complements to ByteTrack, effectively reducing the confusion caused by simple global data association methods. Moreover, our approach demonstrates a substantial advantage over other trackers \cite{MTracker,6, 8, 58, CBIOU} that employ unique designs for data association. We achieve the top ranking in HOTA, IDF1, AssA and DetA. These findings suggest that our data association method itself effectively addresses the challenges posed by camera motion and image blurring in the MOT17 dataset, even without the use of complex components such as motion camera compensation.

\noindent \textbf{MOT20.} The metrics of The performance metrics of DeconfuseTrack on the MOT20 test dataset are presented in \cref{tab:2}. Our method consistently outperforms ByteTrack in all metrics, with a 2\% improvement in HOTA, a 2.4\% improvement in IDF1, a 0.3\% improvement in MOTA, a 3.1\% improvement in AssA, and a 0.7\% improvement in DetA. These findings validate the effectiveness of our proposed enhancements in dense object scenarios. Compared to other trackers, our method ranks first in HOTA, IDF1, and DetA, with MOTA only 0.1\% lower than the top-performing method \cite{46}. Even when compared to methods that utilize reinforcement learning \cite{31}, recurrent neural network \cite{RTU++}, graph neural network \cite{46}, and Transformer \cite{32} in the data association stage, our approach still exhibits a significant advantage. This suggests that through our decomposed association design, strong performance can be achieved with a simple clue extraction and modeling process.

\subsection{Ablation Studies}
\noindent \textbf{Analysis of DDA.} We conduct ablation experiments to validate the effectiveness of the components in DDA, and the results are shown in \cref{tab:3}. Firstly, when using each deconfusion module separately (rows 2-4), we observe significant improvements. Using TDM alone results in a 0.7\% increase in HOTA, 0.6\% increase in MOTA, 1.2\% increase in IDF1, and 1\% increase in AssA. Similarly, using ADM alone leads to a 0.5\% increase in both HOTA and MOTA, a 0.6\% increase in IDF1, and a 1\% increase in AssA. This indicates that the appearance cues effectively alleviate the confusion caused by motion cues. However, when using DDM alone (row 1), the performance gain is relatively low. We speculate that relying solely on low-score detections without utilizing appearance cues limits the ability to deconfuse the motion cues. When combining DDM with TDM (row 5), there is a notable improvement compared to using DDM alone. We believe this is because DDM and TDM synergistically work together, where high-score detections freed by DDM can be further deconfused by TDM after re-association. Finally, when all three sub-modules are used together (row 6), there is an additional 0.4\% increase in MOTA, further demonstrating the effectiveness of DDA.

\begin{table}[t]
\centering
\resizebox{\columnwidth}{!}{

\begin{tabular}{lcccccc}
    \toprule[2pt]
        Method & Venue & HOTA$\uparrow$  & IDF1$\uparrow$  & MOTA$\uparrow$  & AssA$\uparrow$  & DetA$\uparrow$  \\ \midrule
        \multicolumn{6}{l}{Learnable Matcher:}  \\ 
        MeMOT\cite{50} & CVPR'22 & 54.1  & 66.1  & 63.7  & 55.0  & - \\ 
        MOTRv2\cite{32} & CVPR'23 & 60.3  & 72.2  & 76.2  & 58.1  & 62.9 \\ 
        UTM\cite{46} & CVPR'23 & 62.5  & 76.9  & \textbf{78.2}  & - & - \\ 
        \multicolumn{6}{l}{Non-Learnable Matcher:} \\ 
        FairMOT\cite{9} & IJCV'21 & 54.6  & 67.3  & 61.8  & 54.7  & 54.7 \\ 
        RelationTrack\cite{57} & TMM'22 & 56.5  & 70.5  & 67.2  & 56.4  & 56.8 \\ 
        MTracker\cite{MTracker} & ECCV'22 & -  & 67.7  & 66.3 & -  & - \\ 
        \rowcolor{pink!50}ByteTrack\cite{7} & ECCV'22 & 61.3  & 75.2  & {77.8}  & 59.6  & {63.4} \\ 
        \rowcolor{pink!50}QuoVadis\cite{12} & NeurIPS'22 & 61.5  & 75.7  & {77.8}  & 59.9  & {63.3} \\ 
        \rowcolor{pink!50}RTU++\cite{RTU++} & TIP'22 & 62.8  & 76.8  & {76.5}  & 62.6  & {63.1} \\ 
        \rowcolor{pink!50}SAT\cite{31} & ACM MM'22 & {62.6}  & {76.6}  &75.0  & 63.2 & 62.1 \\ 
        \rowcolor{pink!50}StrongSORT++\cite{6} & TMM'23 & 62.6  & 77.0  & 73.8  & \textbf{64.0}  & 61.3 \\ 
        \rowcolor{pink!50}OC-SORT\cite{8} & CVPR'23 & 62.1  & 75.9  & 75.5  & {62.0}  & - \\ 
        \rowcolor{pink!50}GHOST\cite{58} & CVPR'23 & 61.2  & 75.2  & 73.7  & - & - \\ 
        \hline
        \rowcolor{pink!50}\textbf{DeconfuseTrack} &- & \textbf{63.3}  & \textbf{77.6}  & 78.1  & 62.7  & \textbf{64.1} \\ \bottomrule[2pt]
    \end{tabular}}
    \caption{Comparing with state-of-the-art methods on the MOT20 test set under the private detection protocol. The methods within the \sethlcolor{pink!50}\hl{pink} block utilize YOLOX \cite{30} as the detector. The best results are highlighted in \textbf{bold}.}
\vspace{-0.05in}
\label{tab:2}
\end{table}
\begin{table}[t]
\centering
\resizebox{\columnwidth}{!}{%

\begin{tabular}{lccccccc}
\toprule[2pt]
\multicolumn{1}{l}{\multirow{2}{*}{Method}} & \multicolumn{3}{c}{Components} & \multicolumn{4}{c}{Metrics}                                 \\ \cmidrule(r){2-4} \cmidrule(r){5-8} 
\multicolumn{1}{c}{} & DDM    & TDM    & ADM    & HOTA$\uparrow$          & MOTA$\uparrow$ & IDF1$\uparrow$          & AssA$\uparrow$ \\ \hline
Baseline             &     &     &     & 69.0            & 79.4 & 80.6          & 70.6 \\
                     & $\surd$    &     &     & 69.1          & 79.4 & 80.7          & 70.7 \\
                     &     & $\surd$    &     & 69.7          & 80.0   & 81.8          & 71.6 \\
                     &     &     & $\surd$    & 69.5          & 79.9 & 81.2          & 71.6 \\
                     & $\surd$    & $\surd$   &     & \textbf{69.9} & 80.0   & \textbf{82.1} & 71.9 \\
Baseline+DDA                                & $\surd$         & $\surd$         & $\surd$         & \textbf{69.9} & \textbf{80.4} & \textbf{82.1} & \textbf{72.0} \\ \bottomrule[2pt]     
\end{tabular}

}
\caption{
The ablation study of DDA. (DDM: Detection Disambiguation Module, TDM: Trajectory Disambiguation Module, ADM: Association Disambiguation Module)
}
\vspace{-0.05in}
\label{tab:3}
\end{table}

\noindent \textbf{Component-wise Analysis. } In the ablation experiments, we validated the effectiveness of the components proposed in DeconfuseTrack, and the results are shown in \cref{tab:4}. When ONMS was used alone (row 3), there was a 0.6\% increase in MOTA, while the overall tracking performance remained largely unchanged. This is consistent with our hypothesis that ONMS primarily aims to improve the detection stage. When combining DDA and ONMS to form DeconfuseTrack (row 4), significant improvements in various association metrics were observed compared to using DDA alone (row 2). This is because the detections recovered by ONMS can be utilized by the DDM component in DDA, enhancing the accuracy of association in cases of severe occlusion. Overall, compared to the baseline (row 1), DeconfuseTrack achieved a 1.7\% increase in HOTA, a 1.1\% increase in MOTA, a 3.1\% increase in IDF1, and a 3\% increase in AssA, demonstrating notable improvements in both detection and association aspects.
\begin{table}[t]
\centering
\resizebox{0.8\columnwidth}{!}{%

\begin{tabular}{lcccc}
\toprule[1.5pt]
\multicolumn{1}{l}{\multirow{2}{*}{Method}} & \multicolumn{4}{c}{Metrics}                                   \\ \cmidrule{2-5} 
\multicolumn{1}{c}{}                        & HOTA$\uparrow$          & MOTA$\uparrow$          & IDF1$\uparrow$          & AssA$\uparrow$          \\ \midrule
Baseline                                    & 69.0          & 79.4          & 80.6          & 70.6          \\
Baseline+DDA                                & 69.9          & 80.4          & 82.1          & 72.0          \\
Baseline+ONMS                               & 69.1          & 80.0          & 80.7          & 70.6          \\
DeconfuseTrack                              & \textbf{70.7} & \textbf{80.5} & \textbf{83.7} & \textbf{73.6} \\ 
\bottomrule[1.5pt]
\end{tabular}

}
\caption{
Ablation study of the components in DeconfuseTrack. (DDA: Decomposed Data Association, ONMS: Occlusion-aware NMS).
}
\vspace{-0.05in}
\label{tab:4}
\end{table}

\begin{figure}[tp]
    \centering
    \includegraphics[width=\linewidth]{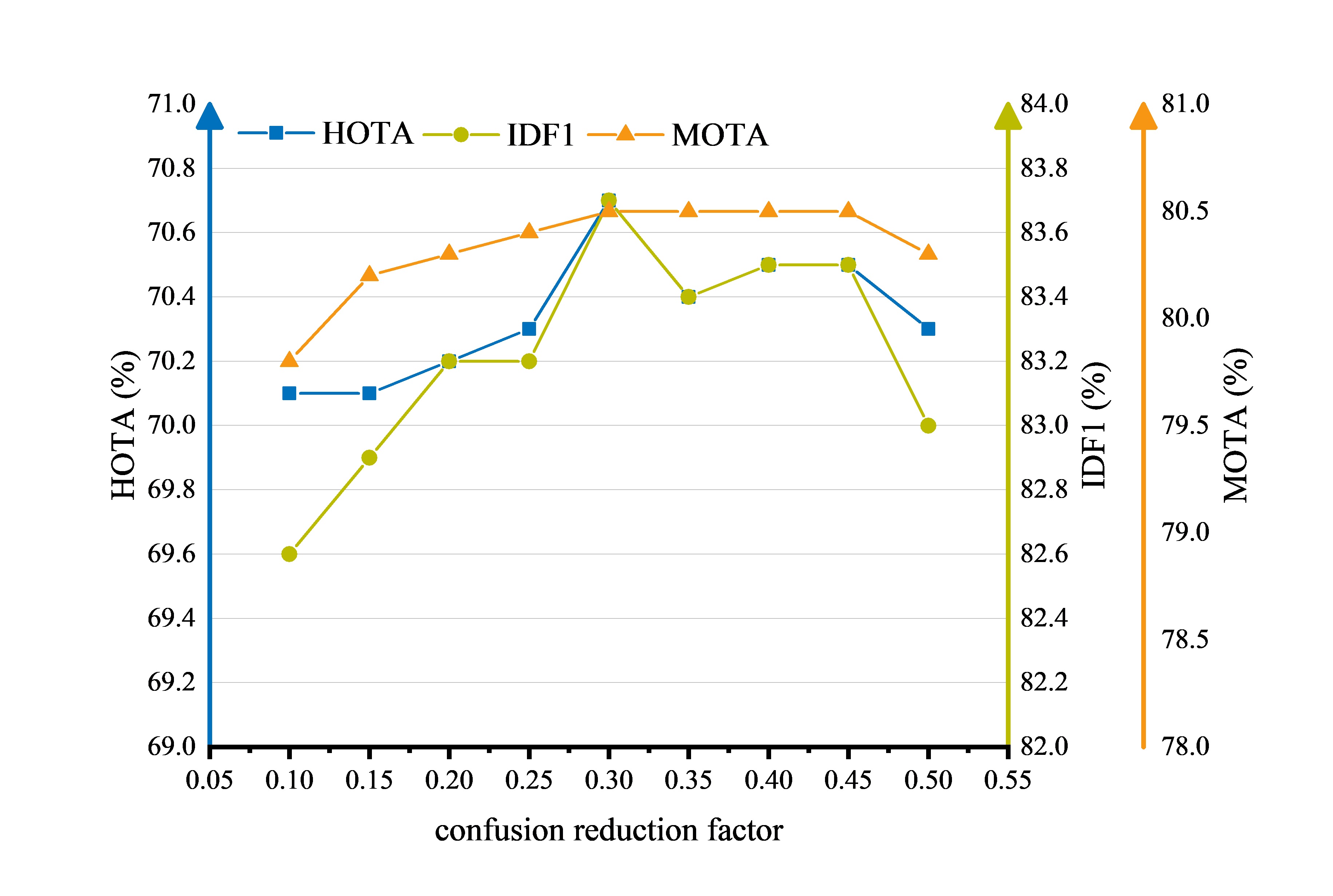}
    \caption{Comparison of the performances of DeconfuseTrack under different detection confusion reduction factor. The results are from the validation set of MOT17.}
    \label{fig:4}
\end{figure}
\noindent \textbf{Robustness to Confusion Reduction Factor. } The confusion reduction factor $\kappa$ is an important hyperparameter in DeconfuseTrack. We adjusted it from 0.1 to 0.5 and compared the tracking metrics. The results are shown in \cref{fig:4}. From the results, we can observe that $\kappa$ is sensitive and achieves the maximum performance when set to 0.3. Therefore, we selected $\kappa$ to be 0.3.

\noindent \textbf{Application on Other Trackers. } We incorporated ONMS and DDA into 5 popular trackers based on the MMTracking\cite{55} framework, and the results are shown in \cref{tab:5}. Our method aims to reduce assignment confusion primarily for trackers that rely on motion cues. As a result, our approach yields good performance on trackers like SORT\cite{4}, Tracktor\cite{34}, and ByteTrack\cite{7}. However, its impact on DeepSORT\cite{5} is not substantial and may even cause a decrease in IDF1. These results indicate that ONMS and DDA have strong generalization capabilities, allowing for easy integration into advanced trackers and obtaining relatively stable performance gains.
\begin{figure}[tp]
    \centering
    \includegraphics[width=\linewidth]{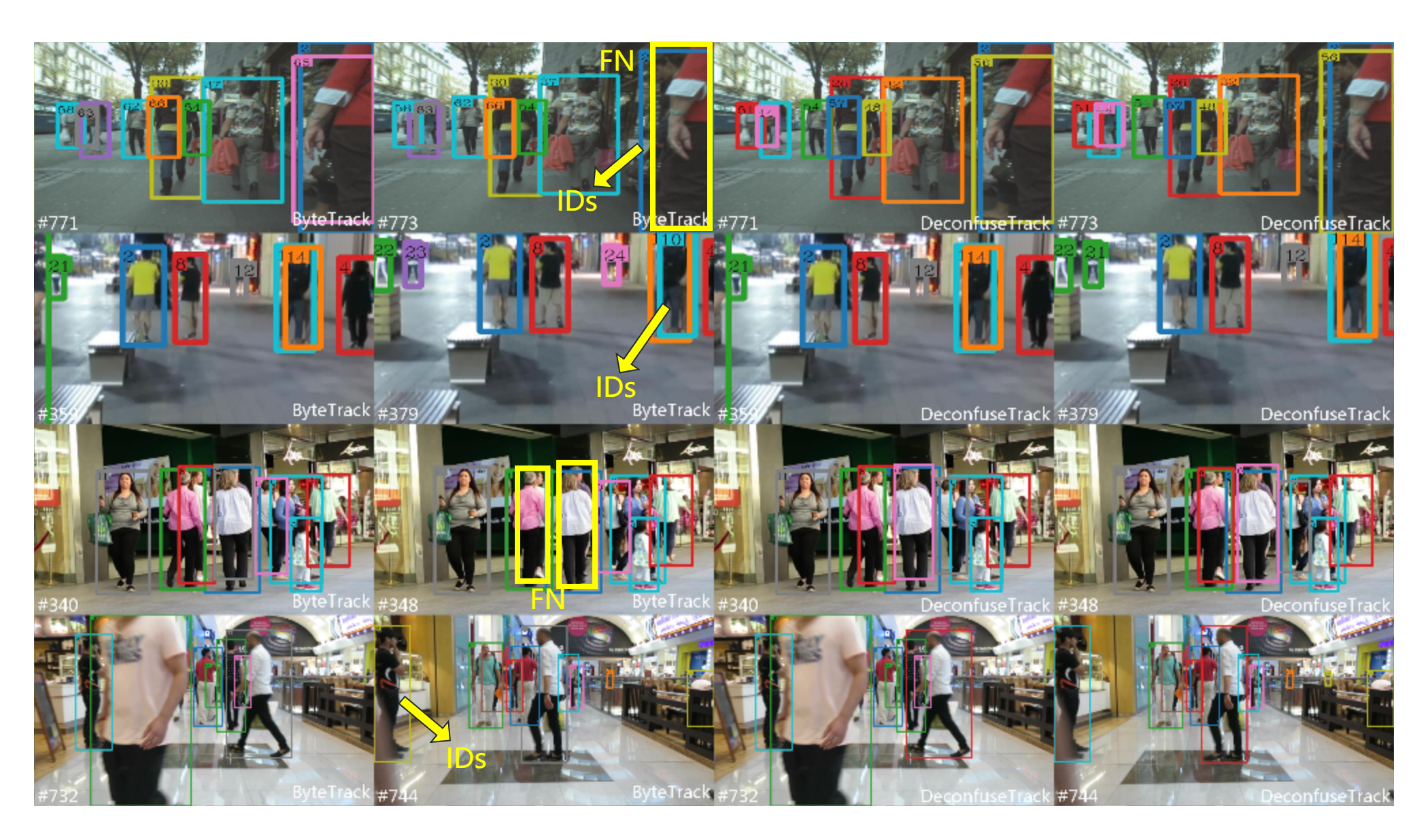}
    \caption{Visualization on the MOT17 validation set. }
    \label{fig:5}
\end{figure}
\begin{table}[t]
\centering
\resizebox{\columnwidth}{!}{%

\begin{tabular}{
lccclllll
}
\toprule[2pt]
\multirow{2}{*}{Method} & \multicolumn{3}{c}{Components} & \multicolumn{4}{c}{Metrics}                                                                               \\ \cmidrule(r){2-4} \cmidrule(r){5-8}
                        & B      & O    & D   & \multicolumn{1}{c}{HOTA$\uparrow$} & \multicolumn{1}{c}{MOTA$\uparrow$} & \multicolumn{1}{c}{IDF1$\uparrow$} & \multicolumn{1}{c}{AssA$\uparrow$} \\ \midrule
\multirow{4}{*}{SORT{\cite{4}}}        &   &   &   & 52.0       & 62.0       & 57.8       & 49.2       \\
                                    & $\surd$ &   &   & 52.9       & 63.0       & 59.7       & 50.5       \\
                                    & $\surd$ & $\surd$ &   & 53.5\textcolor{ForestGreen}{(+0.6)} & 63.8\textcolor{ForestGreen}{(+0.8)} & 61.2\textcolor{ForestGreen}{(+1.5)} & 51.4\textcolor{ForestGreen}{(+0.9)} \\
                                    & $\surd$ & $\surd$ & $\surd$ & 54.0\textcolor{ForestGreen}{(+1.1)} & 65.0\textcolor{ForestGreen}{(+2.0)} & 62.5\textcolor{ForestGreen}{(+2.8)} & 52.1\textcolor{ForestGreen}{(+1.6)} \\ \midrule
\multirow{4}{*}{DeepSORT\cite{5}}    &   &   &   & 57.3       & 63.7       & 69.7       & 59.6       \\
                                    & $\surd$ &   &   & 57.3       & 65.1       & 70.2       & 59.6       \\
                                    & $\surd$ & $\surd$ &   & 57.5\textcolor{ForestGreen}{(+0.2)} & 65.7\textcolor{ForestGreen}{(+0.6)} & 70.0(-0.2) & 59.6(+0.0) \\
                                    & $\surd$ & $\surd$ & $\surd$ & 57.6\textcolor{ForestGreen}{(+0.3)} & 66.1\textcolor{ForestGreen}{(+1.0)} & 70.1(-0.1) & 59.8\textcolor{ForestGreen}{(+0.2)} \\ \midrule
\multirow{4}{*}{Tracktor\cite{34}}   &   &   &   & 52.4       & 61.0       & 59.8       & 51.3       \\
                                    & $\surd$ &   &   & 53.0       & 62.0       & 60.7       & 52.1       \\
                                    & $\surd$ & $\surd$ &   & 53.3\textcolor{ForestGreen}{(+0.3)} & 62.5\textcolor{ForestGreen}{(+0.5)} & 60.8\textcolor{ForestGreen}{(+0.1)} & 52.1(+0.0) \\
                                    & $\surd$ & $\surd$ & $\surd$ & 54.9\textcolor{ForestGreen}{(+1.9)} & 63.5\textcolor{ForestGreen}{(+1.5)} & 64.8\textcolor{ForestGreen}{(+4.1)} & 55.3\textcolor{ForestGreen}{(+3.2)} \\ \midrule
\multirow{4}{*}{Tracktor++\cite{34}} &   &   &   & 55.7       & 64.0       & 66.9       & 57.7       \\
                                    & $\surd$ &   &   & 55.7       & 64.5       & 66.8       & 57.5       \\
                                    & $\surd$ & $\surd$ &   & 56.4\textcolor{ForestGreen}{(+0.7)} & 65.5\textcolor{ForestGreen}{(+1.0)} & 67.6\textcolor{ForestGreen}{(+0.8)} & 57.9\textcolor{ForestGreen}{(+0.4)} \\
                                    & $\surd$ & $\surd$ & $\surd$ & 56.4\textcolor{ForestGreen}{(+0.7)} & 65.6\textcolor{ForestGreen}{(+1.1)} & 67.9\textcolor{ForestGreen}{(+1.1)} & 58.0\textcolor{ForestGreen}{(+0.5)} \\ \midrule
\multirow{4}{*}{OC-SORT\cite{8}}     &   &   &   & 67.8       & 77.4       & 78.0       & 69.3       \\
                                    & $\surd$ &   &   & 68.8       & 79.7       & 79.9       & 70.1       \\
                                    & $\surd$ & $\surd$ &   & 68.9\textcolor{ForestGreen}{(+0.1)} & 79.8\textcolor{ForestGreen!60}{(+0.1)} & 80.2\textcolor{ForestGreen}{(+0.3)} & 70.1(+0.0) \\
                                    & $\surd$ & $\surd$ & $\surd$ & 68.9\textcolor{ForestGreen}{(+0.1)} & 80.0\textcolor{ForestGreen}{(+0.3)} & 80.6\textcolor{ForestGreen}{(+0.7)} & 70.1(+0.0) \\ 
\bottomrule[2pt]
\end{tabular}

}
\caption{
Results of applying ONMS and DDA to popular trackers on the MOT17 validation set. In order to highlight our contribution, we provide results compared to each baseline after adding two-stage data association method BYTE\cite{7}. Performance improvements are indicated in \textcolor{ForestGreen}{green}. (B: BYTE, O: ONMS, D: DDA)
}
\vspace{-0.05in}
\label{tab:5}
\end{table}

\noindent \textbf{Visualization. } We visualized partial results of ByteTrack and our proposed DeconfuseTrack, as shown in \cref{fig:5}. ByteTrack suffers from missed detections and blurry motion clues due to occlusions, image blurriness, camera motion, and other factors. Consequently, it leads to confusion issues such as ID switches, target losses, and localization errors. In contrast, our method incorporates ONMS and DDA, effectively mitigating these problems, which validates the necessity and effectiveness of our deconfusion approach.

\section{Conclusion}
\label{sec:conclusion}
In this study, we propose a novel plug-and-play data association method called DDA and a simple detection post-processing method called ONMS. Based on these methods, we design a tracker named DeconfuseTrack that focuses on addressing confusion issues in MOT. Extensive experiments demonstrate that a more detailed consideration of data association can significantly improve the performance of existing MOT methods. Our work aims to break the current trend of oversimplified data association steps in most MOT methods and provide insights to researchers, inspiring the development of more effective data association techniques. We believe that by carefully considering data association, future MOT methods can achieve higher accuracy and robustness, driving advancements in the field.

{\small
\bibliographystyle{ieeenat_fullname}
\bibliography{11_references}
}


\end{document}